\documentclass[10pt,a4paper,conference]{IEEEtran}
\ifCLASSINFOpdf
\usepackage[pdftex]{graphicx}
\graphicspath{{../pdf/}{../jpeg/}}
\DeclareGraphicsExtensions{.pdf,.jpeg,.png}
\else
\usepackage{graphicx}
\graphicspath{{../eps/}}
\DeclareGraphicsExtensions{.eps}
\fi
\usepackage{threeparttable}
\usepackage{hyperref}

\usepackage{amsmath}
\usepackage{algorithm}
\usepackage{algorithmic}

\usepackage{array}
\usepackage{multirow}

\ifCLASSOPTIONcompsoc
\usepackage[caption=false,font=normalsize,labelfont=sf,textfont=sf]{subfig}
\else
\usepackage[caption=false,font=footnotesize]{subfig}
\fi

\begin{document}
	%
	\title{Fusion of Global-Local Features for Image Quality Inspection of Shipping Label}

	
	
	%
	\author{\IEEEauthorblockN{Sungho Suh\IEEEauthorrefmark{1}\IEEEauthorrefmark{2},
			Paul Lukowicz\IEEEauthorrefmark{2}\IEEEauthorrefmark{3} and
			Yong Oh Lee\IEEEauthorrefmark{1}}
		\IEEEauthorblockA{\IEEEauthorrefmark{1}Smart Convergence Group, Korea Institute of Science and Technology Europe \\Forschungsgesellschaft mbH, 66123 Saarbrücken, Germany}
		\IEEEauthorblockA{\IEEEauthorrefmark{2}Department of Computer Science, TU Kaiserslautern, 67663 Kaiserslautern, Germany}
		\IEEEauthorblockA{\IEEEauthorrefmark{3}German Research Center for Artificial Intelligence (DFKI), 67663 Kaiserslautern, Germany}
		Email: s.suh@kist-europe.de, paul.lukowicz@dfki.de, yongoh.lee@kist-europe.de}


	\maketitle
	
	\begin{abstract}
		
		The demands of automated shipping address recognition and verification have increased to handle a large number of packages and to save costs associated with misdelivery. A previous study proposed a deep learning system where the shipping address is recognized and verified based on a camera image capturing the shipping address and barcode area. Because the system performance depends on the input image quality, inspection of input image quality is necessary for image preprocessing. In this paper, we propose an input image quality verification method combining global and local features. Object detection and scale-invariant feature transform in different feature spaces are developed to extract global and local features from several independent convolutional neural networks. The conditions of shipping label images are classified by fully connected fusion layers with concatenated global and local features. The experimental results regarding real captured and generated images show that the proposed method achieves better performance than other methods. These results are expected to improve the shipping address recognition and verification system by applying different image preprocessing steps based on the classified conditions.
		
	\end{abstract}
	
	
	\IEEEpeerreviewmaketitle
	
	\section{Introduction}
	
	With the rapid rise of the logistics delivery market, the annual cost associated with failed delivery has also increased. In 2015, this annual cost was estimated to be approximately 1.5 billion dollars for the postal service and 20 billion dollars for the mailing industry in United States of America \cite{UAACost:2015}. One of the main causes of misdelivery is incorrect shipping labels, which may contain falsely given addresses or may be damaged during the logistics process. Packages with problematic labels lead to not only cost loss during unnecessary delivery to undeliverables and return but also damage to the manufacturer’s and seller’s reputations. A process for screening undeliverable addresses and error-containing barcodes in the shipping label that is used for identifying packages and their destinations is required.
	
	To verify the information in the shipping label, the commonly used approaches are optical character verification (OCV) and optical character recognition (OCR). These approaches capture an image using a vision system or smartphone, detect regions of interest (ROIs), and recognize optical characters in the ROIs. Several traditional image processing methodologies have been applied for OCV and OCR, but deep-learning-based approaches have been shown to be effective for these tasks \cite{zhang2015symmetry, gupta2016synthetic, shi2017end, EAST:CVPR17, baek2019character}. Previous studies adopting a deep neural network framework provided high accuracy of verification and recognition of expiration dates in a food package or address in a shipping label \cite{Ribeiro:ICIP18, suh2019robust, thota2020multi}. 
	
\begin{figure*}[!t]
		\centering
		\subfloat[Normal]{\includegraphics[height=3cm]{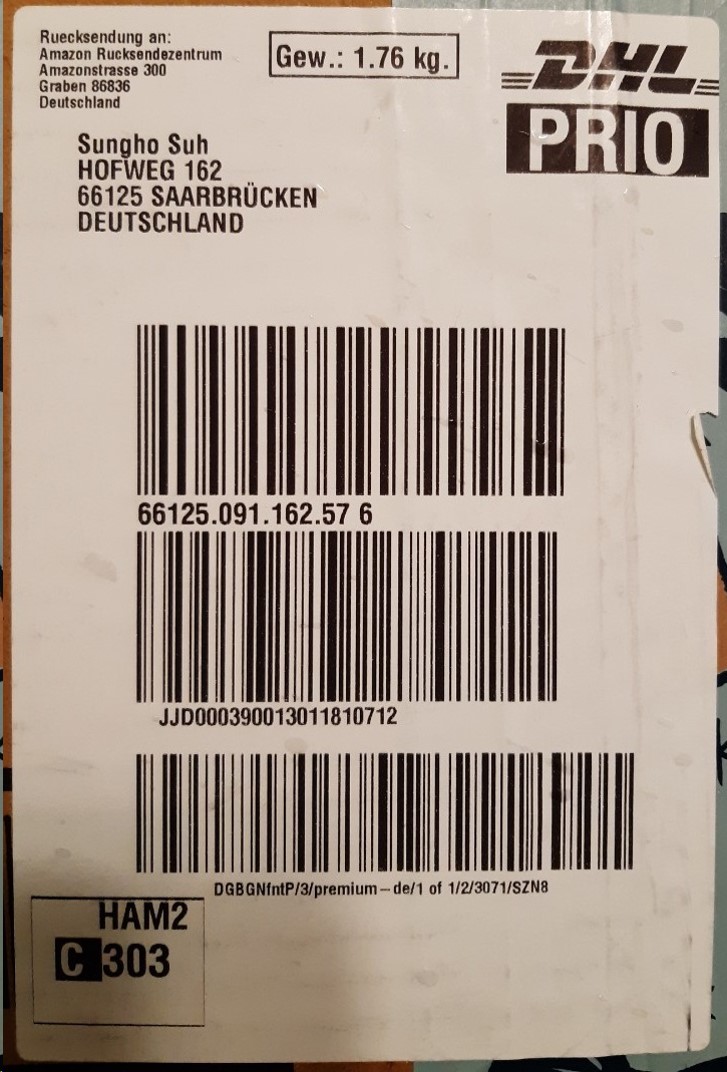}}
		\hfil
		\subfloat[Contaminated]{\includegraphics[height=3cm]{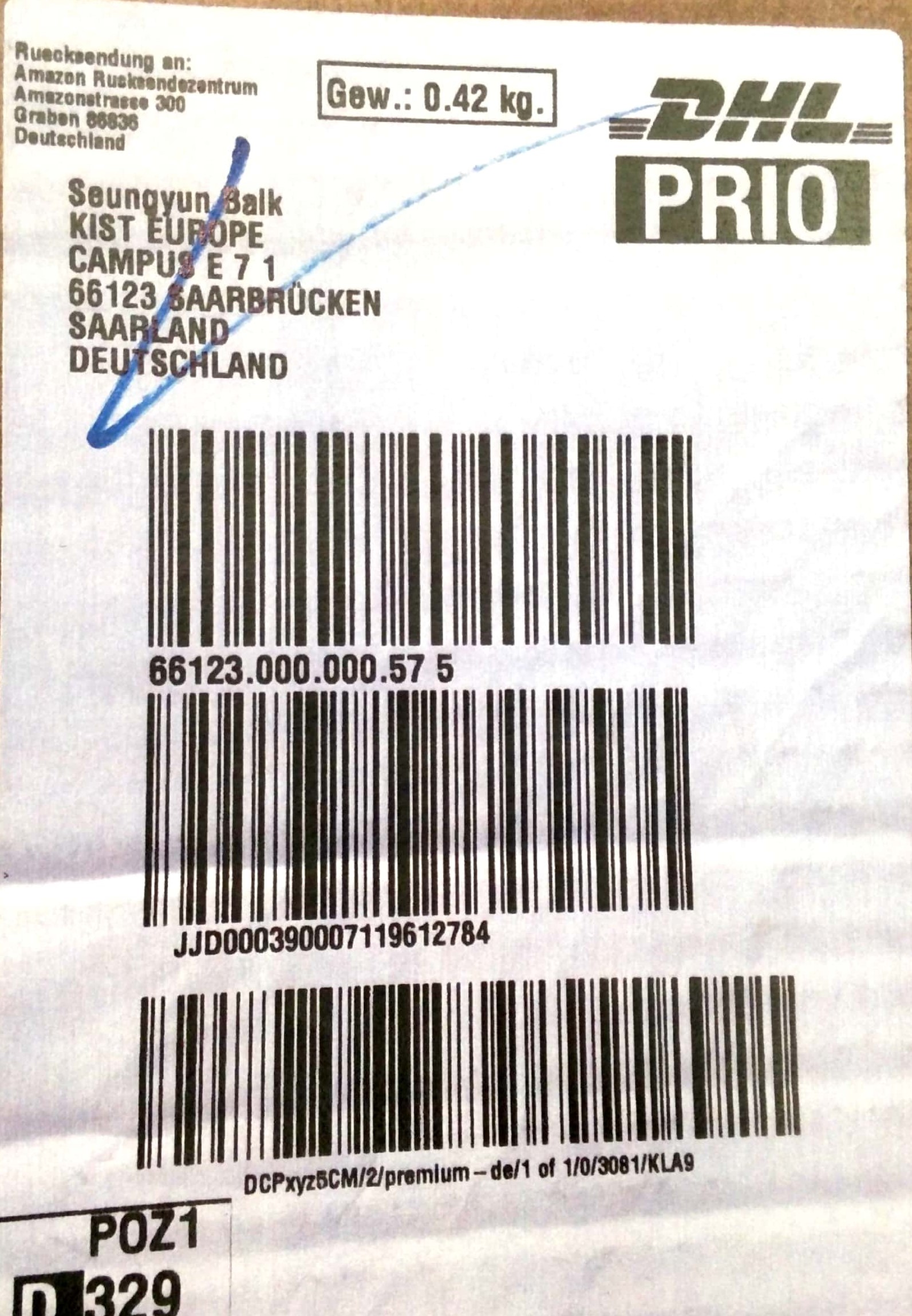}}
		\hfil
		\subfloat[Unreadable]{\includegraphics[height=3cm]{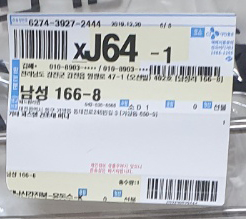}}
		\hfil
		\subfloat[Handwritten]{\includegraphics[height=3cm]{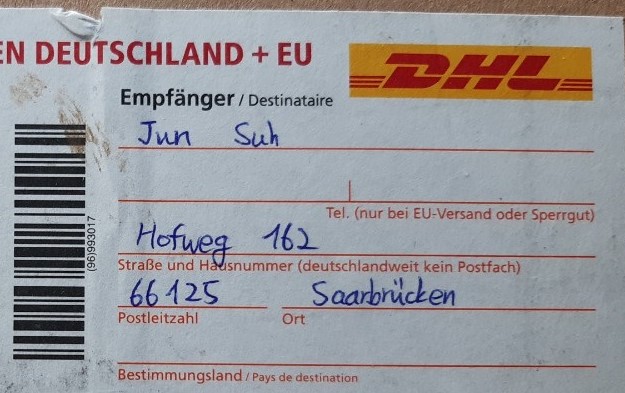}}
		\hfil
		\subfloat[Damaged]{\includegraphics[height=3cm]{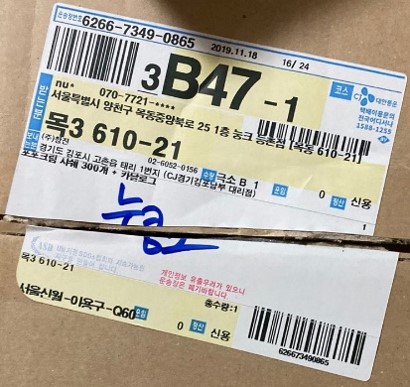}}
		\caption{Examples of the annotated real dataset of the shipping label.}
		\label{example_shippinglabel}
	\end{figure*}

\begin{figure*}[!t]
		\centering
		\subfloat[Normal]{\includegraphics[height=3cm]{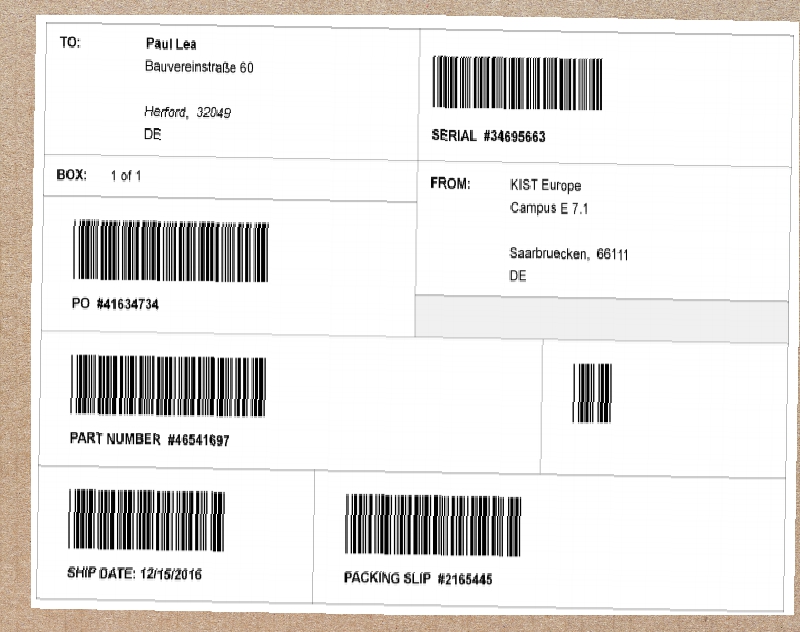}}
		\hfil
		\subfloat[Contaminated]{\includegraphics[height=3cm]{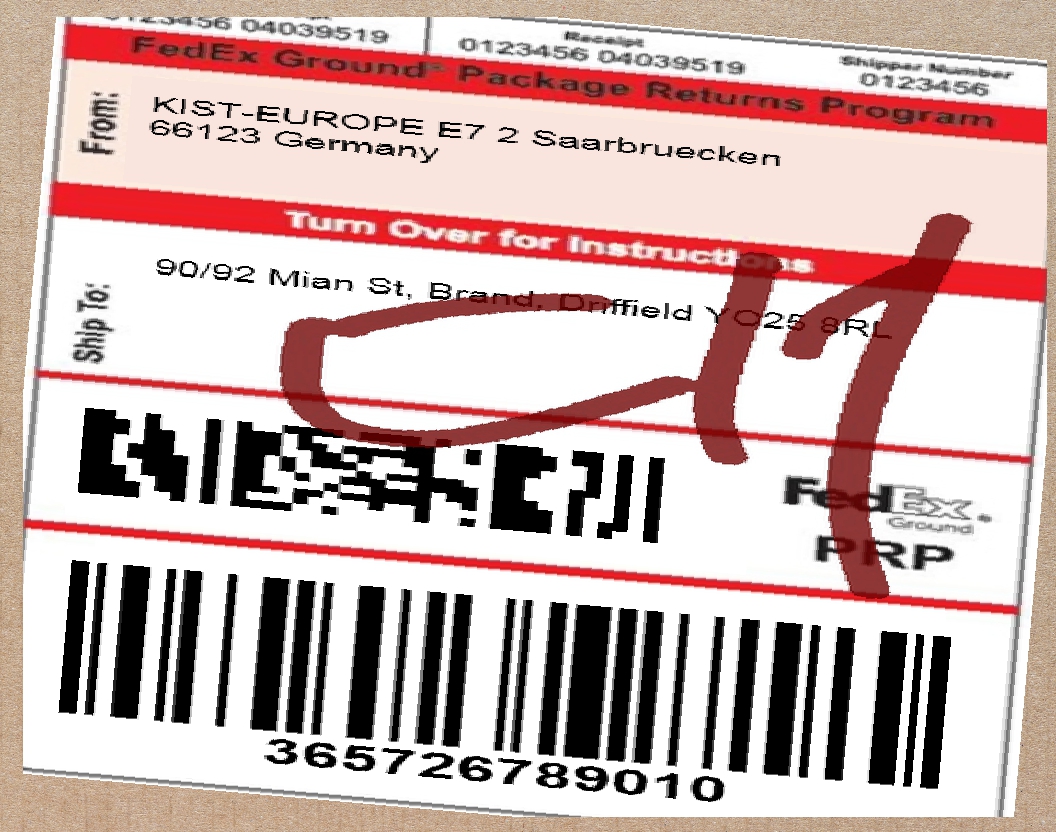}}
		\hfil
		\subfloat[Unreadable]{\includegraphics[height=3cm]{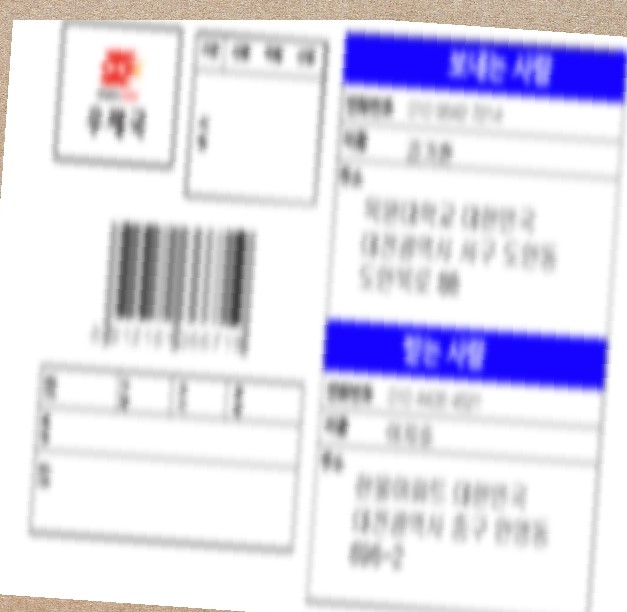}}
		\hfil
		\subfloat[Handwritten]{\includegraphics[height=3cm]{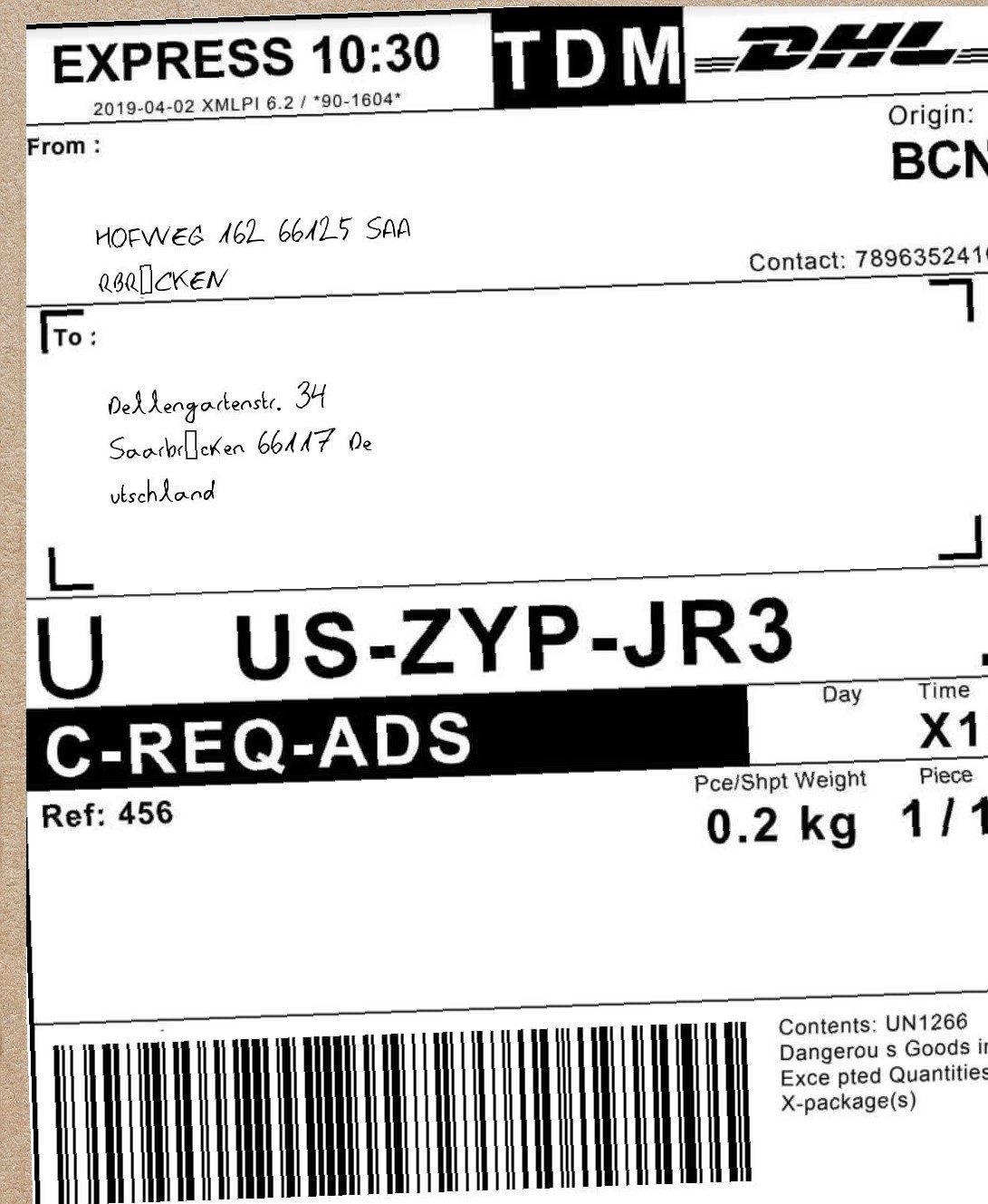}}
		\hfil
		\subfloat[Damaged]{\includegraphics[height=3cm]{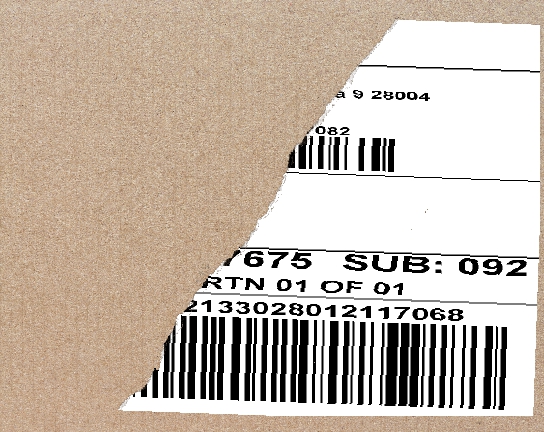}}
		\caption{Examples of the annotated generated dataset of the shipping label.}
		\label{example_generatedshippinglabel}
	\end{figure*}

	However, the performance of OCR engines and text detection engines is sensitive to image quality and defects on target objects. The quality of the captured image can be affected by many degradations caused during image acquisition. Further, during the packaging and delivery processes, the shipping label can be damaged. Thus, proper assessment and classification of the captured image and object are required to improve the text detection and recognition processes. Four poor conditions of captured images are defined, and example images of generated and collected datasets are shown in Fig. \ref{example_shippinglabel} and Fig. \ref{example_generatedshippinglabel}.
	
	\begin{enumerate}[]
		\item Unreadable: Blurring is the most common issue in real-world applications, occurring because of defocusing or camera motion \cite{li2018cg}, and an image acquired at the wrong positioning leads to missing the ROI. By detecting the blur degradation and incorrect positioning, the user reacquires the image and the OCR accuracy can be improved. 
		\item Contaminated: Shipping labels are often contaminated in the delivery process, or some part of the address area in the shipping label can be occluded by a pen mark. By processing with methods such as document image enhancement and binarization \cite{he2019deepotsu, vo2018binarization}, the address in the contaminated or occluded label can be recognized better. 
		\item Handwritten: The shipping label sometimes contains handwritten addresses. Because recognition engines for machine-printed characters and handwritten characters are different, the two types of characters must be classified differently.
		\item Damaged: The shipping label can be torn or occluded by another shipping label in the processes of packaging and delivery. By detecting damaged shipping labels, misdelivery can be reduced and the performance of the text recognition process can be improved.
	\end{enumerate}
	
	In this paper, we propose an input image quality verification method using convolutional neural networks (CNNs) for shipping label inspection. The proposed method classifies the mentioned previously five image types: normal, contaminated address, unreadable image, handwritten address, and damaged shipping label. 
	
	However, classification methods based on CNNs are not suitable for direct use for input image quality verification because of two problems. One problem is the varying aspect ratios and sizes of shipping label images. The varying aspect ratios make it difficult to design a CNN with a fixed image size as input. Another problem is that it is difficult to distinguish between contaminants in the address area and contaminants in other areas. In general, there are two methods for handling input images of different sizes: (1) resizing the input image while maintaining the aspect ratio in the padding space and (2) using patches as inputs to the CNN to extract discriminative features. However, resizing the input images leads to increasing the ambiguity between normal and blurred images because of the loss of detailed features, making it difficult to detect defects in ROIs. Conversely, dividing an input image into patches can lead to the loss of global features, degrading the classification performance.
	
	We integrate local and global features from different CNNs. To localize the local features, we detect the barcode and address areas using the deep-learning object detection algorithm You Only Look Once (YOLO) \cite{YOLO:CVPR16, yolov3}. Then, we localize and crop regions with strong features using one of the famous feature extraction methods, features from accelerated segment test (FAST) \cite{rosten2006machine}. The global features are extracted from the global CNNs with a resized input image, and the local features are extracted from independent local CNNs with the detected address and barcode area and the localized regions using FAST. We concatenate the global and local features together into fully connected fusion layers to classify the shipping label image conditions. The contribution of this paper can be summarized into two points. (1) We detect and localize the ROIs using YOLO and FAST and extract the local features using independent local CNNs. (2) By constructing the variant global and local images to combine the global and local features, we propose a novel CNN model-based stacked generalization ensemble method \cite{brownlee2018better} for image quality verification.
	
	The rest of the paper is organized as follows. Section \ref{sec:relatedwork} summarizes related works. Section \ref{sec:method} describes the details of the proposed algorithm, and Section \ref{sec:experimentalresults} presents the experimental results. Finally, Section \ref{sec:conclusion} concludes the paper with discussion of the results.
	
	\section{Related Work}
	\label{sec:relatedwork}
	Many inspection methods for printed labels have been studied \cite{tamura2010development, grosso2011automated, sarkar2015image, pedersen2016quality}. A previous study \cite{suh2019robust} proposed a deep-learning system where the shipping address is recognized and verified based on a camera image capturing the shipping address and barcode area. As mentioned in the previous section, proper assessment and classification of the captured image and object are required to improve the text detection and recognition processes. Several traditional image-processing methodologies, including stroke width transform \cite{epshtein2010detecting}, scale-invariant feature transform \cite{xie2013fast}, and maximally stable extremal region \cite{chen2011robust}, have been applied to detect text regions and inspect the printing quality. However, deep-learning-based approaches have been shown to be especially effective in these tasks \cite{Ribeiro:ICIP18, suh2019robust, thota2020multi}. Ribeiro et al. \cite{Ribeiro:ICIP18} proposed automatic recognition of expiration dates in food packages using EAST \cite{EAST:CVPR17}. This method evaluated food package image quality using CNNs and the localized date region by fully connected networks and EAST. However, the ambiguity between normal and blurred images is aggravated, as is the classification of defects in ROI and defects outside of ROI because the image quality evaluation using CNNs used only the resized entire image.
	
	Several papers have proposed CNN-based classification methods that integrate global and local features. Tian et al. \cite{tian2018selective} proposed a vehicle model recognition model using an iterative discrimination CNN based on selective multi-convolutional region feature extraction. The SMCR features consisted of global and local features, and the classification model combined global and local features. Lu et al. \cite{lu2019integrating} presented a script identification method integrating local and global CNNs in natural scene images. They trained the global CNN with segmented images and the local CNN with patches. He et al. \cite{he2019traffic} proposed a traffic sign recognition method by combining global and local features, which were extracted by histograms of oriented gradient (HOG), color histograms (CH), and edge feature (EF). These methods showed that the CNN-based classification combining global and local features provided better performance than the CNN-based classification with only an entire image.

	\section{Proposed Method}
	\label{sec:method}
	
\begin{figure}[!t]
		\centering
		\includegraphics[width=\columnwidth]{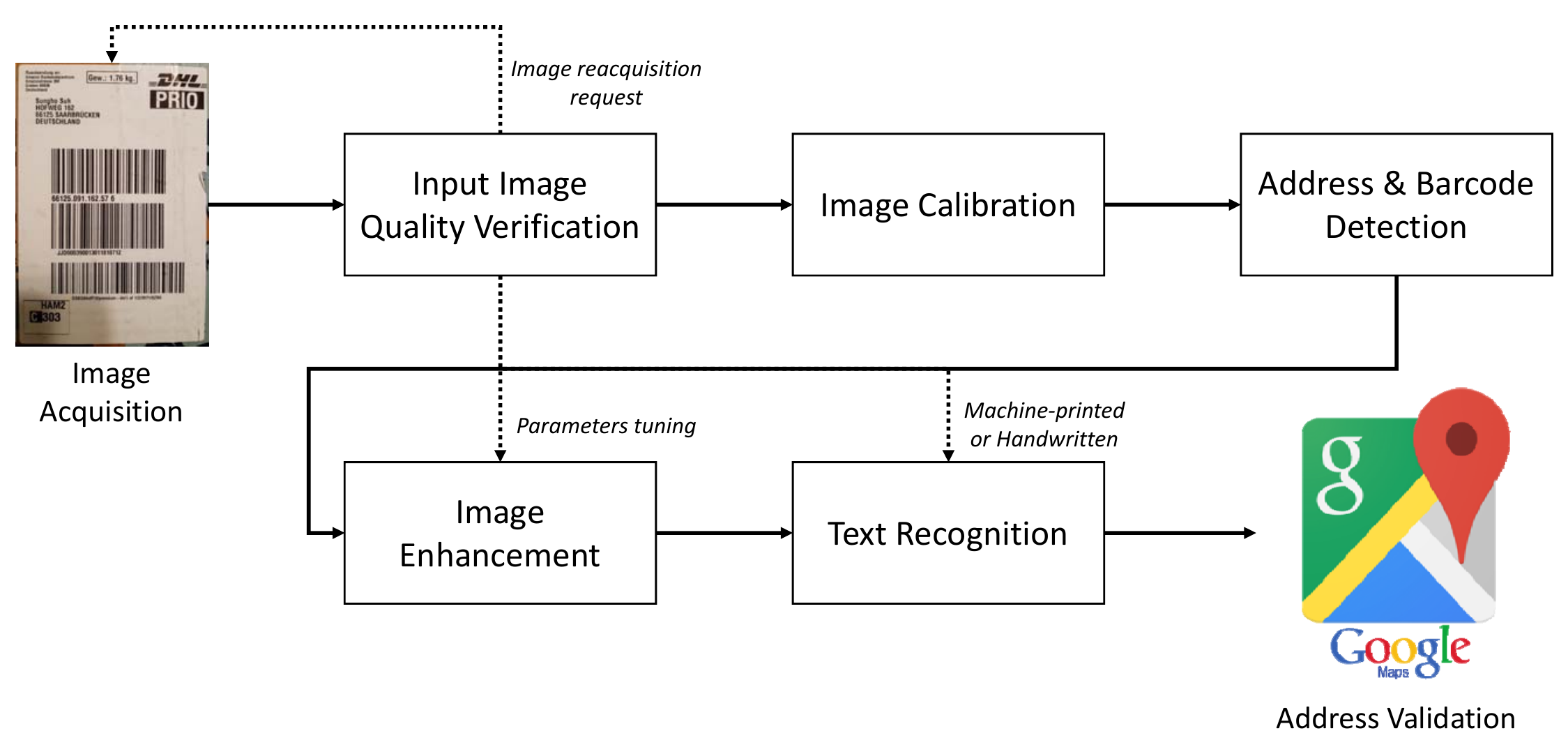}
		\caption{Flowchart of the shipping label inspection system}
		\label{fig_overview}
	\end{figure}

	The shipping label inspection system \cite{suh2019robust} involves six steps: input image quality verification, calibration, salient region detection, image enhancement, text recognition, and address validation. The flowchart of the shipping label inspection system is demonstrated in Fig. \ref{fig_overview}. 
	
	In the first step, the proposed input image quality verification process classifies the five image types: normal, contaminated address, unreadable image, handwritten address, and damaged shipping label. The unreadable image and the damaged label are reported to the user to acquire the image properly or check the condition of the shipping label. The contaminated and handwritten addresses provide information on the address area for image enhancement and/or text recognition processes. 
	
\begin{figure*}[!t]
		\centering
		\includegraphics[width=2\columnwidth]{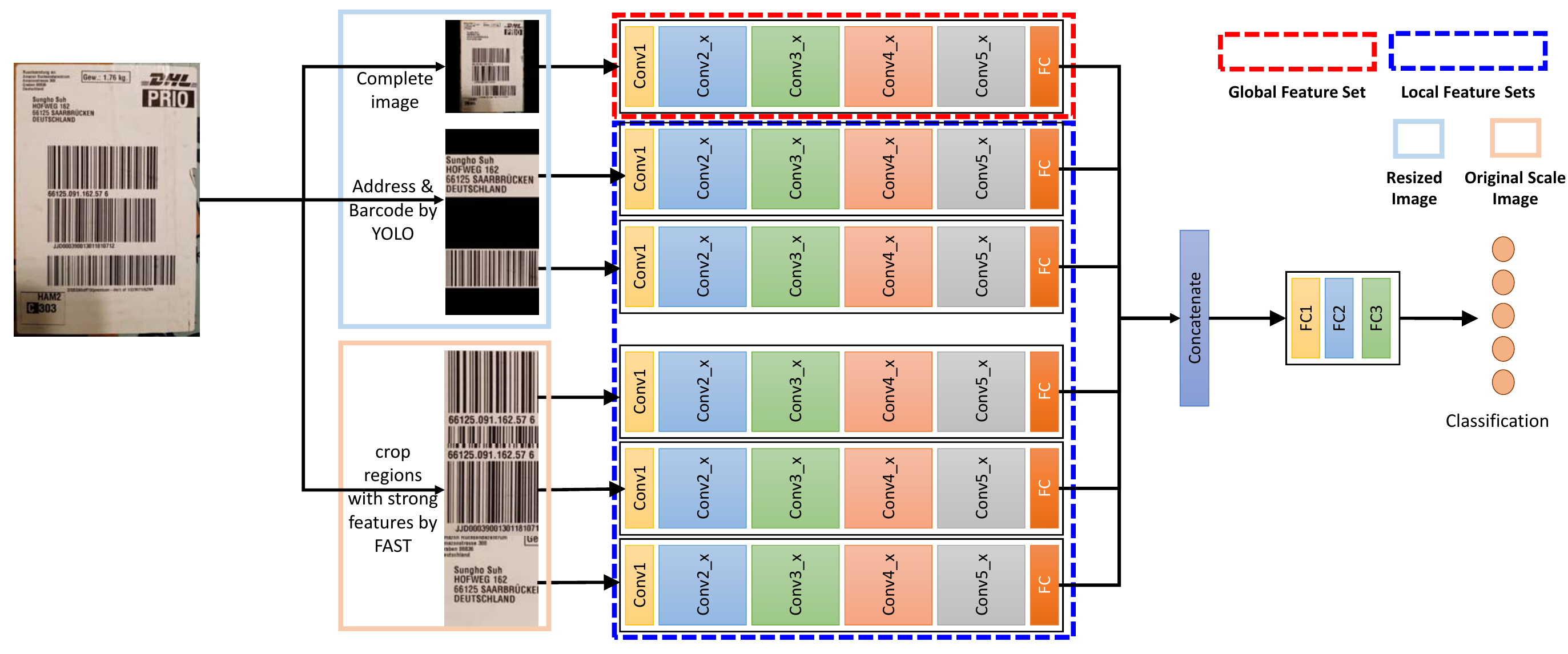}
		\caption{Overall network architecture of the proposed method}
		\label{fig_proposed_architecture}
	\end{figure*}

	The goal of the input image quality verification method is to classify the five image types. We propose an input image quality verification method using convolutional networks with global and local features. The overall network architecture of the proposed method is presented in Fig. \ref{fig_proposed_architecture}.
	
	\subsection{Feature Localization}
	
	Classification with a resized entire input image leads to ambiguity between normal and blurred images because of the loss of detailed features, making it difficult to detect contaminated defects and damaged defects in ROIs. For these reasons, we detect barcode and address areas and crop regions with strong features to extract local features. In this study, we train global and local CNNs to extract the global and local features and combine the global and local features, as shown in Fig. \ref{fig_proposed_architecture}.
	
	\begin{algorithm}[H]
		\caption{Feature localization using the FAST algorithm. We use the default threshold value of the FAST algorithm, $t=50$, patch size $D_p=(256,256)$, and number of patches $n_p=3$}\label{FASTAlgo}
		\begin{algorithmic}[1]
			\renewcommand{\algorithmicrequire}{\textbf{Input:}}
			\renewcommand{\algorithmicensure}{\textbf{Output:}}
			\REQUIRE input image of shipping label
			\ENSURE localized regions using FAST
			\STATE Convert input image to grayscale image
			\STATE Detect feature points $p$ using FAST with $t$
			\STATE $D_i=$size of input image
			\IF {($D_i>D_p$)}
			\STATE $patchNum_i=D_i/D_p$
			\STATE $L_p=$ number of $p$
			\FOR {each $j\in [1, patchNum_i]$}
			\STATE $iCountPoint = 0$
			\STATE $R_j=$ region of $j$-th patch
			\FOR {each $k\in [1, L_p]$}
			\IF {($p(k) \in R_j$)}
			\STATE $iCountPoint = iCountPoint + 1$
			\ENDIF
			\ENDFOR
			\STATE $iCountFeature(j) = iCountPoint$
			\ENDFOR
			\STATE Extract the $n_p$ largest patches and crop the region $I_P$
			\ELSE
			\STATE Copy the image to $I_P$ in the padding space of size $D_p$
			\ENDIF 
			\RETURN $I_P$
		\end{algorithmic}
	\end{algorithm}

	To detect and localize the ROIs, we use YOLO to detect barcode and address areas and FAST to localize regions with strong features. YOLO is a fast general-purpose object detector using deep neural networks that can achieve real-time detection~\cite{yolov3}, which is much faster than R-CNN-~\cite{ashraf2016shallow} and Faster R-CNN~\cite{ren2015faster}-based methods. Unlike conventional region-based object detection algorithms, the YOLO detector considers object detection as an end-to-end regression problem and uses a single convolutional network to predict the bounding boxes and the corresponding class probabilities. We fine-tune the YOLO detector to locate the barcode and address areas in the shipping label images. We exploit the YOLO network implemented on the Darknet19~\cite{darknet13} framework with a resized input image of size $416\times416$. After detecting the barcode and address regions, we crop and resize the detected regions to $256\times256$ while maintaining the aspect ratio in the padding space to set the input in the local CNNs.
	
	The global features are extracted from the global CNNs with a resized input image, and the local features are extracted from independent local CNNs with the detected address and barcode area, and the localized regions using FAST. FAST is a famous corner-detection algorithm that can be used to extract feature points; it demonstrates rapid operation and a low number of computations compared with other feature-detection methods. We calculate the number of feature points from the FAST algorithm in a patch of size $256\times256$, the same as the input image size in the local CNNs. By choosing patches with a large number of feature points, regions with strong features, such as barcodes or characters, can be selected as input images of local CNNs. Additionally, because the patches are not resized, but rather maintain the original scale of the input image, they can provide helpful information to classify unreadable input images using the local CNNs.
	
	\subsection{Global and Local Feature Fusion}
	Global features describe the overall outline of the input image, whereas local features focus on the local shape, the degree of contamination in the localized regions, lighting conditions, and other imaging factors. To extract the global and local features, a resized complete input image is processed in the global CNNs, and the localized images are processed in the independent local CNNs, as shown in Fig. \ref{fig_proposed_architecture}. In this study, we deploy ResNet-50 \cite{he2016deep} pre-trained on ImageNet \cite{russakovsky2015imagenet} for global and local feature extractions. In the proposed architecture, four independent CNNs are trained for feature extraction: a global CNN, two local CNNs with address and barcode areas by YOLO, and a local CNN with regions cropped by FAST. The local CNNs with regions cropped by FAST share the network parameters in the training procedure. 
	
	To combine the global and local features, we adopt a stacked generalization ensemble. Stacked generalization is an ensemble method where a new model learns how to combine the predictions from multiple models \cite{brownlee2018better}. To control the importance of the feature combination, we add a fully connected layer with different output feature dimensions to the feature extraction architecture. In the proposed model, we set the dimension of the global feature to 2048 and the dimension of each local feature to 512. To verify the proposed model, we compare the proposed ensemble models with majority voting and weighted majority voting algorithms \cite{kolter2007dynamic}.

	\section{Experimental Results}
	\label{sec:experimentalresults}
	In this section, we verify the proposed method on our generated and collected real dataset and compare the results with other image classification methods: ResNet-50 \cite{he2016deep} and VGGNet-19 \cite{simonyan2014very}. The experiments were all implemented using Python scripts in the Pytorch framework \cite{paszke2017automatic} and tested on a Linux system. Training procedures were performed on NVIDIA Tesla V100 GPUs.
	
	\subsection{Dataset}
	
	To the best of our knowledge, there is no open database for shipping labels because of privacy issues. Because of the scarcity of shipping label images, we generated and collected 5306 and 1092 images of different types and from various countries using smartphones. Note that we collected real data with the agreement of the owner, and we plan to release the dataset onto our public website. The dataset includes images of the shipping labels, the position information of the barcode and address, and five annotated labels: normal, contaminated, unreadable, handwritten, and damaged. Five people manually annotated the dataset with specific sorting criteria. In the case of the ``unreadable'' class, the receiver's address is blurred or distorted, or the resolution is too low to recognize the address. For the ``contaminated'' and handwritten class, the receiver's address is contaminated, occluded by pen mark, or handwritten. Finally, a damaged shipping label indicates that the shipping label is torn or occluded by another shipping label. Detailed information about the dataset is presented in Table \ref{table:shippinglabeldataset}.
	
	\begin{table}[!t]
		\caption{Manually annotated dataset of the shipping labels}
		\label{table:shippinglabeldataset}
		\centering
		\begin{tabular}{|c|c|c|}
			\hline
			Defect Type	& \# of Images (Generated)  & \# of Images (Collected) \\ \hline
			Normal &    1283  &   660\\ 
			Contaminated  &    1054  & 139  \\ 
			Unreadable & 904 & 107 \\ 
			Handwritten & 988 & 52\\ 
			Damaged & 1077 & 134\\ \hline
			Total & 5306 & 1092\\ \hline
		\end{tabular} 
	\end{table}
	To further evaluate the performance of the proposed method, we obtained the shipping label formats from 15 different carrier companies, generated barcodes randomly, and entered random addresses from seven different countries. The label images were generated in various sizes using NiceLabel Designer software \cite{nicelabel}. To evaluate the proposed method, we generated contaminated, unreadable, and damaged images using image-processing methods, and then we generated handwritten labels using Google handwriting fonts \cite{Googlefont}. Finally, there were 5306 generated images for input image quality verification. We selected 100 images per class as the test set and conducted a 10-fold cross-validation procedure for the dataset.
	
	The collected dataset contains only 1092 images and is under a data-imbalanced condition, while the generated dataset has 5306 images and the number of data for each class is similar. To improve the performance of the classification on the collected real dataset, a data augmentation method, such as rotation (90, 180, 270 degrees) and flip, is adopted to extend the data of minority classes. To evaluate the proposed method using deep neural networks, we conducted a 10-fold cross-validation procedure for the shipping label dataset.
	
	\subsection{Experimental Results on Generated Images}
	
	\begin{table}[!t]
		\caption{Detection accuracy by YOLO}
		\label{table:segmentationresults}
		\centering
		\begin{tabular}{|c|c|c|}
			\hline
			Region  & mAP (Generated) & mAP (Collected)\\ \hline
			Address &    0.9653 $\pm$ 0.1637     & 0.8907 $\pm$ 0.0215 \\ 
			Barcode  &    0.9976 $\pm$ 0.0018    &    0.9778 $\pm$ 0.0065\\ \hline
			Total & 0.9814 $\pm$ 0.0085 & 0.9343 $\pm$ 0.0105\\ \hline
		\end{tabular} 
	\end{table}
	
	\begin{figure}[!t]
		\centering
		\subfloat[]{\includegraphics[width=\columnwidth]{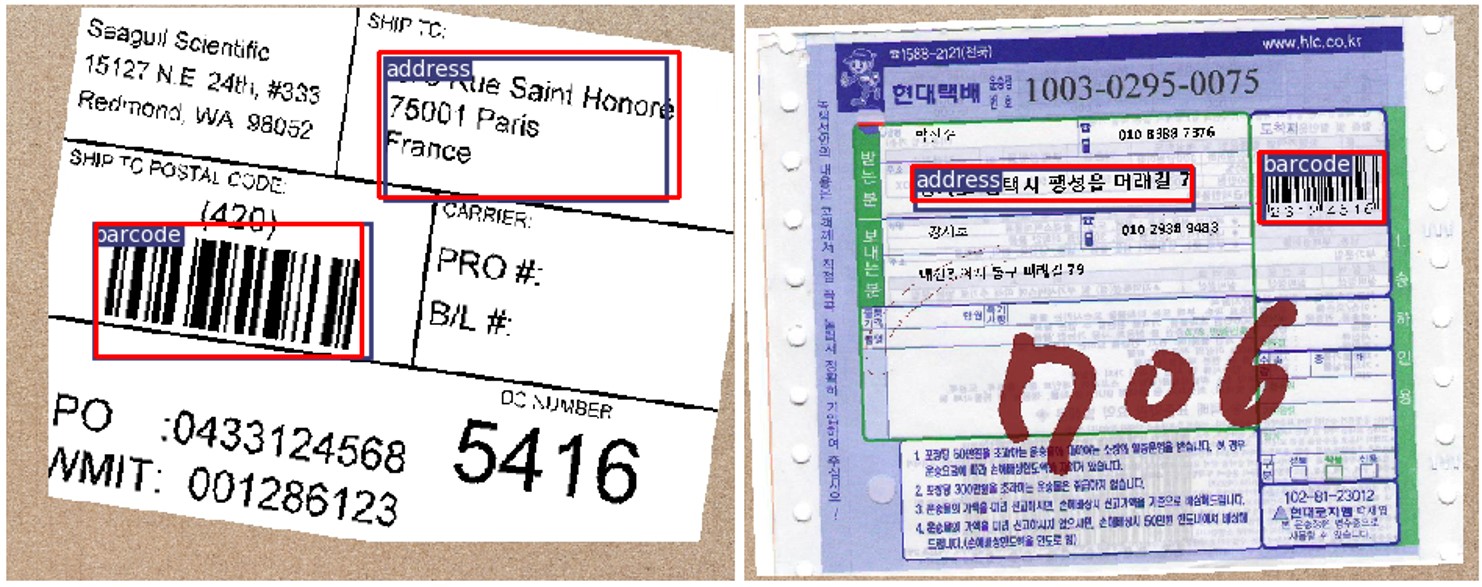}}
		\hfil
		\subfloat[]{\includegraphics[width=\columnwidth]{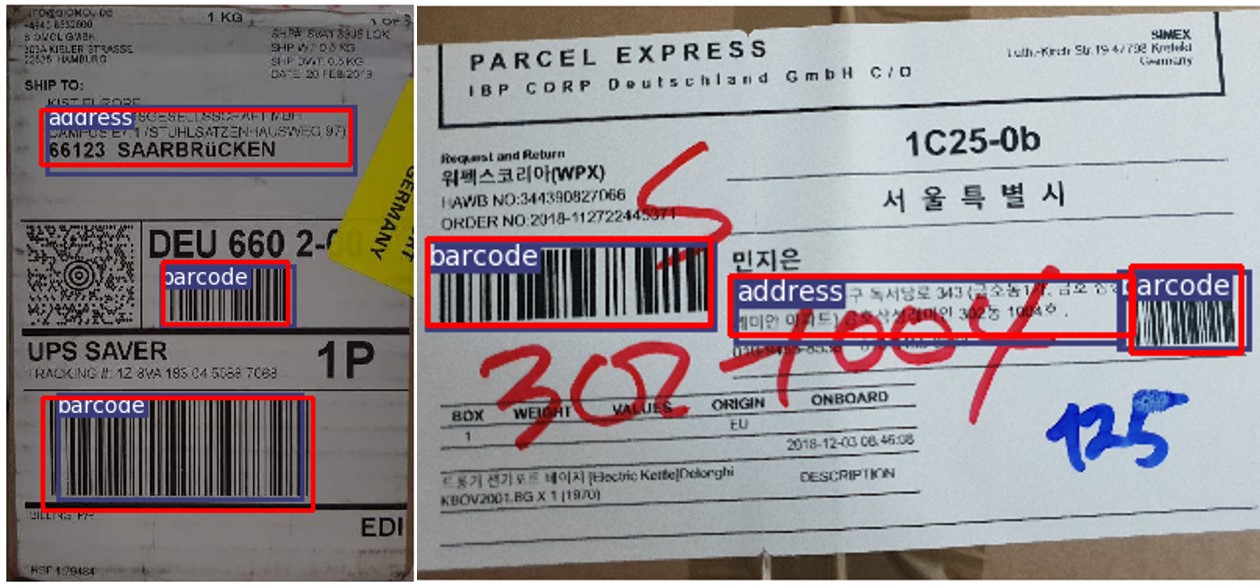}}
		\caption{Examples of detected barcode and address areas (a) on generated images and (b) on the real dataset: red boxes are ground truth, and blue boxes are detected regions}
		\label{example_YOLO}
	\end{figure}
	
	\begin{table*}[!t]
		\renewcommand{\arraystretch}{1.3}
		\caption{Comparison of classification results on the generated real dataset}
		\label{table:generateddatasetresult}
		\centering
		\begin{tabular}{|c|c|c|}
			\hline
			Methods & VGG-19 Classification Accuracy & Resnet-50 Classification Accuracy \\
			\hline
			Only global features & 95.98 $\pm$ 0.74 \% & 95.80 $\pm$ 0.38 \%\\
			Global-local fusion(majority voting) &  96.40 $\pm$  0.65\% &  96.62 $\pm$  0.43\%\\
			Global-Local fusion (weighted majority voting) &  97.16 $\pm$  0.43 \% & 97.02 $\pm$ 0.77 \%\\		
			\hline
			Global-Local fusion (ours) & 98.32 $\pm$ 0.49\% & \textbf{ 99.06 $\pm$ 0.66\%}\\
			\hline
		\end{tabular}
	\end{table*}
	
	We evaluate the proposed method on the generated dataset in this subsection. Table \ref{table:segmentationresults} shows the performance of the feature localization by YOLO. The results are the mean average precision \cite{Liu2009Learning} and show mean and standard deviation. The feature localization on the generated dataset provided a high performance, above 0.98, and the example images of the detection results are shown in Fig \ref{example_YOLO} (a).  
	
	For the evaluation, VGG-19 and ResNet-50 were used for classification, and we compared the proposed method by combining the global and local features of the classification methods with only global features and other ensemble learning methods, such as majority voting and weighted majority voting. Table \ref{table:generateddatasetresult} shows the classification accuracy comparison with respect to the generated dataset. The proposed method with ResNet-50 provided the highest classification accuracy of 98.46 \%, which exceeded ResNet-50 with only global features and other ensemble learning methods by 3.46\% and 2.04\%, respectively. 
	
	\begin{table*}[!t]
		\renewcommand{\arraystretch}{1.3}
		\caption{Comparison of classification results on the collected real dataset}
		\label{table:realdatasetresult}
		\centering
		\begin{tabular}{|c|c|c|}
			\hline
			Methods & VGG-19 Classification Accuracy & Resnet-50 Classification Accuracy \\
			\hline
			Only global features & 85.40 $\pm$ 2.43 \% & 86.00 $\pm$ 3.40 \%\\
			Global-local fusion(majority voting) &  84.67 $\pm$ 2.27 \% &  86.40 $\pm$ 2.97 \%\\
			Global-Local fusion (weighted majority voting) &  85.00 $\pm$ 2.68 \% & 87.60 $\pm$ 2.70 \%\\		
			\hline
			Global-Local fusion (ours) & 87.80 $\pm$ 2.13\% & \textbf{89.26 $\pm$ 2.70\%}\\
			\hline
		\end{tabular}
	\end{table*}

	\begin{table}[!t]
		\caption{Accuracy of each class on the collected real dataset and comparison with the proposed method with ResNet-50 with only global features}
		\label{table:classresult}
		\centering
		\begin{tabular}{|c|c|c|}
			\hline
			Class  & Only global features & Global-local fusion (ours)\\ \hline
			Normal &    93.33 \%     & 95.00 \% \\ 
			Contaminated  &    78.33 \%    &    85.00 \% \\
			Unreadable &    83.33 \%    &    88.33 \% \\
			Handwritten &    98.33 \%    &    98.33 \% \\
			Damaged & 76.67 \%  & 79.31 \% \\ \hline
			Total & 86.00 \% & 89.26 \% \\ \hline
		\end{tabular} 
	\end{table}

	\subsection{Experimental Results on Collected Real Images}

	We evaluate the proposed method on collected real datasets compared with other classification models with only global features and other ensemble learning methods. Compared with the generated dataset, the collected real dataset has more image types, and it is more difficult to extract local features and classify the input image quality. Table \ref{table:segmentationresults} shows the performance of the feature localization by YOLO. The performance of the feature localization on the collected real images is 0.0471 lesser than the performance on the generated images. Example images of the detection results are shown in Fig \ref{example_YOLO} (b). 
	
	Because the collected real dataset is under an imbalanced data condition, we select 30 images per class as the test set and conduct a 10-fold cross-validation procedure for the dataset. Table \ref{table:realdatasetresult} presents the classification accuracy comparison on the collected real images. It can be observed that ResNet-50 has better performance than VGGNet-19, and the classification with ensemble learning methods also provided higher accuracy than the classification with only global features. This result means that the feature localization method that we proposed improves the classification of the input image quality. In all the methods in Table \ref{table:realdatasetresult}, our method with ResNet-50 achieved the highest classification accuracy of 89.26 \%, combining the global and local features by the stacked generalization method. It improved the classification accuracy by 3.26 \% and 2.86 \% compared with ResNet-50 with only global features and other ensemble learning methods. Unlike other ensemble learning methods, the proposed method concatenates the global and local features and classifies them via three fully connected layers while maintaining the global and local features. However, the classification accuracy on the collected real dataset is lesser than that on the generated dataset. The collected real dataset has more image types and more serious defects. Furthermore, unlike the generated dataset, in which an image is matched one-on-one with a type of defect class, the collected real dataset may have a case where a single image corresponds to a variety of defect types. For example, when a damaged image is occluded by a pen mark and acquired with blurred focus, the image has three types of defects. 
		
	Table \ref{table:classresult} shows the accuracy of each class on the collected real dataset and improvement of the classification by the global-local fusion method. Owing to the detection of address and barcode areas by YOLO, the classification accuracy for the contaminated class was improved by 6.7\%. In addition, the classification accuracy for the unreadable class was improved by 5.0\% by the local features extracted by the FAST algorithm and by maintaining the original scale.
	
	\section{Conclusion}
	\label{sec:conclusion}
	In this paper, we have presented an input image quality verification method using CNNs combining global and local features for shipping label inspection. We detected and localized the regions of interest using YOLO and FAST, and we extracted the global and local features using several independent CNNs. To combine the global and local features, we adopted a stacked generalization ensemble. To train and test the proposed method, we generated and collected a shipping label dataset. The experimental results show that the proposed method improved the classification accuracy by 3.46\% and 3.86\%, respectively, compared with the classification model with only global features, and by 2.04\% and 2.86\%, respectively, compared with the other ensemble learning method combining global and local features. In addition, we plan to release the generated and collected real dataset onto our public website. As future work, the proposed method will be applied to a packaging machine with an industrial camera and will be tested in the logistics industry. We will also improve the image enhancement and text recognition process by applying the proposed method.
	
	\section{Acknowledgment}
	This research was supported by the KIST Europe Institutional Program (Project No. 12005).

	\bibliographystyle{IEEEtran}
	\bibliography{refs}

\end{document}